\documentclass[10pt,twocolumn,letterpaper]{article}
\usepackage{cvpr}              
\usepackage[accsupp]{axessibility}

\newcommand{\TODO}[1]{\textbf{\color{red}[TODO: #1]}}
\renewcommand{\TODO}[1]{}

\usepackage{booktabs,pifont}
\usepackage{siunitx}

\definecolor{cvprblue}{rgb}{0.21,0.49,0.74}
\usepackage[breaklinks,colorlinks,allcolors=cvprblue]{hyperref}
\usepackage{caption}
\usepackage{stfloats}
\usepackage{cuted}
\usepackage{booktabs}
\usepackage{multirow}
\usepackage{graphicx}
\usepackage[table]{xcolor} 
\usepackage{array}         
\usepackage[utf8]{inputenc}
\usepackage{textgreek}
\usepackage{tabularx}   
\usepackage{amsmath,amssymb,mathtools}

\newcolumntype{V}{!{\color{gray!55}\vrule width 0.45pt}} 
\renewcommand{\arraystretch}{1.05}
\newcolumntype{d}[1]{S[table-format=#1]}
\renewcommand{\arraystretch}{1.12}
\newcolumntype{Y}{>{\raggedright\arraybackslash}X}
\newcolumntype{G}{@{\hspace{1.1em}}}
\newcommand{\sdrow}{\rowcolor{gray!3}\scriptsize}

\def\confName{CVPR}
\def\confYear{2026}

\title{EMMA: Extracting Multiple physical parameters 
from Multimodal Data}

\author{
  Farhat Shaikh
  \qquad
  Ayan Banerjee
  \qquad
  Sandeep Gupta\\
  IMPACT Lab, School of Computing 
  \& Augmented Intelligence (SCAI) \\
  Arizona State University, 
  Tempe, AZ\\
  {\tt\small \{fshaik12, abanerj3, 
  Sandeep.Gupta\}@asu.edu}}

\begin{document}
\maketitle


\begin{abstract}

We introduce EMMA, a physics-informed multimodal 
framework that recovers all identifiable dynamical 
parameters of a system directly from raw video, 
audio, and image-based time-series observations. 
Unlike prior video-only approaches that struggle 
with occluded states, hidden actuation inputs, 
or assumptions about known initial conditions 
and coordinate frames, EMMA performs joint 
inference of explicit parameters, implicit 
dynamical components, and calibration invariants 
within a unified continuous-time model. 
EMMA leverages a Liquid Time-Constant (LTC) 
network to learn latent dynamics from 
heterogeneous modalities while a 
physics-constrained loss enforces consistency 
with the governing differential equations. 
A unified feature pipeline enables consistent 
alignment across video trajectories, acoustic 
signatures, and chart-derived measurements, 
allowing EMMA to estimate parameters under 
forced, implicit, and multivariate dynamics 
without requiring segmentation masks, 
differentiable rendering, or specialized sensors. 
Across 100+ scenarios including five standard 
dynamical benchmarks (75 Delfys videos), 
real-world rover and quadrotor systems with 
hidden inputs, and simulation-chart case studies 
spanning biological and chaotic systems, 
EMMA delivers robust multi-parameter recovery 
and significantly outperforms existing 
single-modality and equation-discovery baselines. 
Our results establish EMMA as a general, 
scalable solution for physics-consistent 
model extraction from opportunistic 
multimodal data. Code and data are available at: \url{https://github.com/ImpactLabASU/EMMA-CVPR2026}
\end{abstract}

\begin{figure*}[t]
\centering
\includegraphics[width=0.99\textwidth]{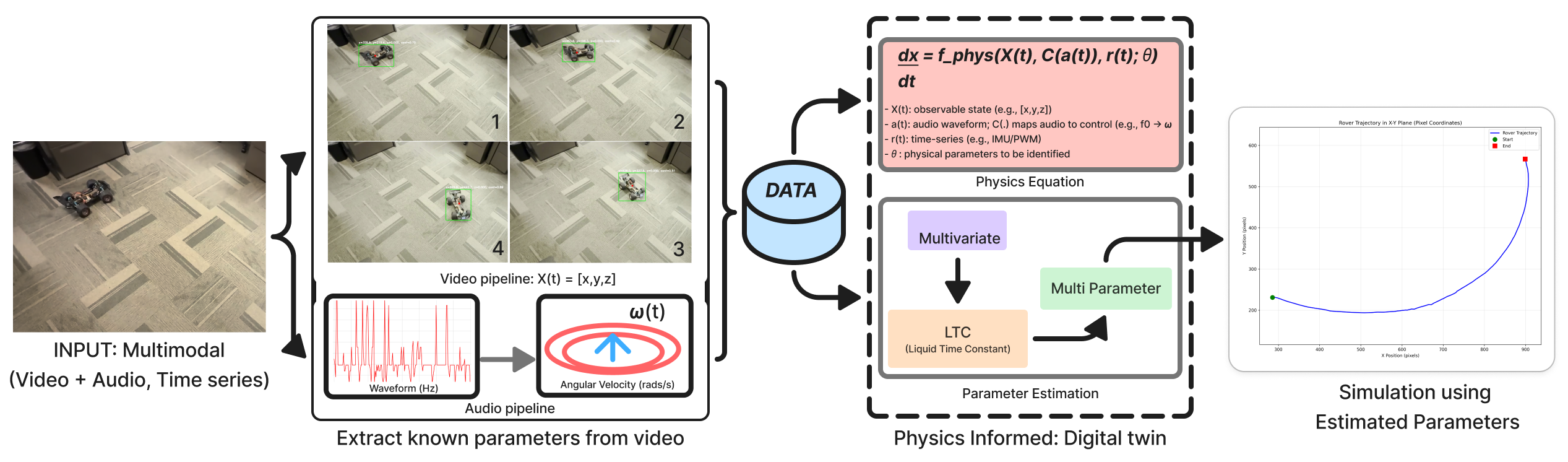}
\caption{Given unified multi-modal observations 
(video, audio, image), \textbf{EMMA} extracts 
identifiable physical parameters through a 
physics-informed digital twin. The Liquid 
Time-Constant network learns dynamics in 
latent space while physics-constrained loss 
enforces consistency with governing equations. 
The estimated parameters enable accurate 
forward simulation without requiring frame 
reconstruction or segmentation masks.}
\label{fig:teaser1}
\end{figure*}


\section{Introduction}
\label{sec:introduction}
Learning the dynamical parameters that govern 
real-world physical systems directly from 
multi-modal vision data is essential for 
constructing high-fidelity digital twins 
of autonomous platforms, such as drones 
and planetary rovers, for testing, simulation, 
fault diagnosis, and safety-critical 
decision support~\cite{Banerjee25Jetc,banik2023dtavs,
wang2024dtad}. 
This task is an instance of inverse modeling, 
where latent physical parameters must be 
inferred from observable 
trajectories~\cite{tarantola2005inverse,
kaipio2006inverse}.

Recent efforts have explored estimating 
dynamical parameters solely from 
video~\cite{hofherr2023neural,
asenov2020vid2param,
kandukuri2020learning,
kandukuri2022ijcv,
ma2022risp}, 
motivated by the fact that classical inverse 
modeling requires accurate measurements of 
all state variables, often necessitating 
intrusive or expensive onboard 
sensors. Vision-only approaches seek to 
bypass this requirement by inferring states 
from passive, widely available 
cameras~\cite{murartal2017orbslam2}. 
However, we observe that video alone 
frequently occludes key state variables, 
especially when systems operate under 
unobserved forcing inputs. For instance, 
in a rover excursion sequence, video frames 
reveal wheel pose but not wheel-power 
commands, making correct kinematic inference 
ill-posed. Audio, in contrast, may encode 
these hidden inputs e.g., wheel rotation 
acoustics strongly correlate with motor speed, 
highlighting the need for multi-modal 
(audio-visual) dynamical parameter 
estimation~\cite{bulatovic2022melspec,
bulatovic2022improve}.

Even with multi-modal inputs, certain 
contributors to system behavior, such as 
frictional drag or terrain-dependent resistive 
forces remain unmeasured. These correspond to 
implicit dynamics, i.e., latent components 
of the governing equations that cannot be 
directly sensed but may still be observable 
through nonlinear dependencies among measured 
states~\cite{chen2016dobc,
nazari2015uio,kaheman2020sindy}. 
Recovering such implicit terms is crucial 
for building physically faithful digital 
twins~\cite{boschetti2024robotics, banerjee2024emily, banerjee2024ecai}, yet 
remains largely unaddressed in existing 
vision-based equation-discovery 
pipelines~\cite{garcia2025lfv,
kandukuri2022ijcv,hofherr2023neural}.

Another challenge in extracting dynamical 
models from video is that many existing 
methods implicitly assume access to invariants 
such as initial conditions, coordinate-system 
origins, or fixed reference 
frames~\cite{ma2022risp,
kandukuri2022ijcv,
kairanda2022phisft}. 
These assumptions rarely hold in real-world 
video, where camera pose, scene geometry, 
and the absolute coordinate origin are unknown. 
Therefore, a practical model-extraction system 
must not only recover the governing dynamical 
parameters but also calibrate these invariants 
jointly, ensuring that the recovered model 
is expressed in the correct physical 
coordinate frame.

As summarized in Table~\ref{tab:relWorks}, 
state of the art (SOTA) video-based dynamical 
parameter estimation methods typically 
(i) ignore external forcing inputs, 
(ii) recover only a single or limited subset 
of parameters, 
(iii) fail to handle implicit dynamical 
components, and 
(iv) assume access to known invariants such 
as initial conditions or coordinate origins.

In this paper, we introduce \textbf{EMMA}, 
a unified multi-modal audio-visual framework 
for dynamical parameter extraction that 
addresses all four limitations identified 
earlier. As shown in Fig.~\ref{fig:teaser1}, 
EMMA ingests synchronized video, audio, and 
\textit{auxiliary timeseries signals extracted 
from visual charts and 
figures}~\cite{rohatgi2017wpd}. 
The video pipeline estimates observable states 
$X(t)$ (e.g., $[x,y,z]$), while the audio 
pipeline recovers latent actuation inputs 
such as wheel angular velocity $\omega(t)$ 
from acoustic signatures, resolving state 
occlusions created by unobserved forcing 
inputs. These signals are fused inside a 
physics-informed parameter estimator built 
on multivariate Liquid Time-Constant (LTC) 
dynamics, which simultaneously 
(1) recovers explicit physical parameters, 
(2) infers \textit{implicit dynamical 
components} that do not appear directly in 
any sensing modality but influence motion 
through nonlinear interactions, and 
(3) performs \textit{invariant calibration} 
by estimating unknown coordinate-frame origins 
and initial conditions directly from raw video. 
The recovered parameters are then used to 
simulate system trajectories, validating 
physical consistency as shown on the right 
side of Fig.~\ref{fig:teaser1}. Together, 
these components enable robust model recovery 
in realistic, instrumentation-limited 
autonomous-system settings.


\paragraph{Our main contributions are summarized below:}
\begin{itemize}
    \item \textbf{Multi-modal dynamical 
    parameter extraction:} A unified 
    framework that estimates multiple 
    dynamical parameters from video, audio, 
    and \textit{time-series reconstructed 
    from visual charts and figures}, as 
    visualized in Fig.~\ref{fig:teaser1}.
    \item \textbf{Recovery under unobserved 
    forcing inputs:} A method for inferring 
    latent actuation inputs such as wheel 
    speed from audio when these inputs are 
    occluded in video, enabling parameter 
    estimation under hidden forcing.
    \item \textbf{Estimation of implicit 
    dynamics:} A mechanism for identifying 
    parameters of unmeasured or latent 
    physical effects (e.g., frictional drag) 
    that shape system behavior but are not 
    directly observable in any modality.
    \item \textbf{Invariant calibration from 
    raw video:} Joint estimation of dynamical 
    parameters and coordinate-system 
    invariants, eliminating assumptions about 
    known initial conditions, camera world 
    alignment, or fixed reference frames.
    \item \textbf{Extensive experimental 
    validation:} Evaluation on the Delfys 
    video benchmark (75 videos), new audio, 
    visual rover and drone datasets, and 
    parameter extraction from 
    \textit{visual charts}, establishing 
    EMMA as a general-purpose multi-modal 
    inverse modeling framework.
\end{itemize}
\vspace{-6pt}

\section{Related work}
\label{sec:related_work}

Table \ref{tab:relWorks} shows a survey of 
all competing works that extract physical 
parameters from video.
\textbf{Baselines.} We adopt \textbf{Delfys} 
\cite{garcia2025lfv} as the primary baseline 
because it most closely matches our problem 
setting: \emph{unsupervised, video} recovery 
of physical parameters for known 
continuous-time ODEs. It is 
\emph{decoder-free} avoiding frame prediction 
and therefore remains stable on real videos 
with intensity or scale variations giving 
good accuracy. Competing video physics methods 
differ substantially in scope: some rely on 
\emph{differentiable rendering with known 
geometry or templates} (gradSim/rSim, 
$\phi$-SfT) 
\cite{jatavallabhula2021gradsim,
kairanda2022phisft}, 
others focus on \emph{state or action 
estimation} rather than parameter 
identification (RISP) \cite{ma2022risp}, 
or operate under \emph{different supervision 
regimes} such as simulation trained models 
(Vid2Param) or setups with measured control 
inputs~\cite{asenov2020vid2param,
kandukuri2022ijcv}. 
Approaches like NIRPI and PAIG address simpler, 
unforced systems with a small number of 
unknowns \cite{hofherr2023neural,
jaques2020physics}. 
For completeness, we also report general 
equation discovery frameworks 
PySINDy/SINDy-PI and PINNs which require 
a video to state or video to field front-end 
and are thus not 
video-native \cite{brunton2016discovering,
desilva2020pysindy,
kaheman2020sindy,
raissi2019physics}. 
Consequently, our head to head comparisons 
emphasize \textbf{Delfys} as the most 
faithful and directly comparable baseline 
to EMMA.

\begin{table*}[t]
\centering
\resizebox{\textwidth}{!}{
\begin{tabular}{l|c|c|c|c|c|c|c}
\toprule
\toprule
\textbf{Work} 
& \textbf{Forcing input} 
& \textbf{Multi-parameter} 
& \textbf{Multivariate} 
& \textbf{Implicit dynamics} 
& \textbf{Multi-modal} 
& \textbf{Learnable Invariant} 
& \textbf{Comparators} \\
 & \textbf{(external $u(t)$)} 
 & \textbf{($>$3 unknowns)} 
 & \textbf{dynamics} &  & \textbf{input} 
 & \textbf{knowledge} & \\
\midrule
\midrule
Delfys \cite{garcia2025lfv} 
  & \ding{55} & \ding{55} & \checkmark 
  & \ding{55} & \ding{55} & \ding{55} 
  & Yes: Primary quantitative baseline. \\
NIRPI \cite{hofherr2023neural} 
  & \ding{55} & \ding{55} & \checkmark 
  & \ding{55} & \ding{55} & \ding{55} 
  & Yes: Quantitatively compared in results. \\
PAIG \cite{jaques2020physics} 
  & \checkmark & \ding{55} & \checkmark 
  & \ding{55} & \ding{55} & \ding{55} 
  & Yes: Quantitatively compared in results. \\
gradSim / rSim \cite{jatavallabhula2021gradsim} 
  & \checkmark & \checkmark & \checkmark 
  & \ding{55} & \ding{55} & \ding{55} 
  & No \\
RISP \cite{ma2022risp} 
  & \checkmark & \ding{55} & \checkmark 
  & \ding{55} & \ding{55} & \ding{55} 
  & No \\
$\phi$-SfT \cite{kairanda2022phisft} 
  & \checkmark & \checkmark & \checkmark 
  & \ding{55} & \ding{55} & \ding{55} 
  & No \\
Vid2Param \cite{asenov2020vid2param} 
  & \ding{55} & \ding{55} & \checkmark 
  & \ding{55} & \ding{55} & \ding{55} 
  & No \\
Kandukuri \cite{kandukuri2022ijcv} 
  & \checkmark & \ding{55} & \checkmark 
  & \ding{55} & \ding{55} & \ding{55} 
  & No \\
\midrule
EMMA ours 
  & \checkmark & \checkmark & \checkmark 
  & \checkmark & \checkmark & \checkmark \\
\bottomrule
\end{tabular}}
\caption{Comparison of related work on model 
recovery \emph{from video}. \checkmark\ = 
demonstrated; \ding{55}\ = not demonstrated.}
\label{tab:relWorks}
\end{table*}


\section{Method}
\label{sec:method}

In this section, we describe components 
of EMMA in detail.

\subsection{Problem Formulation}
\label{sec:problem}

\noindent\textbf{Unified multi-modal inputs.} 
The input to EMMA can be video 
$\{I_t\}_{t=1}^T$ with 
$I_t\!\in\!\mathbb{R}^{3\times H\times W}$ 
or audio $\{A_t\}_{t=1}^T$ with 
$A_t\!\in\!\mathbb{R}^{L_a}$, or images 
of charts $\{M\}$ with 
$M\!\in\!\mathbb{R}^{3 \times H_m \times W_m}$ 
from sensors or cameras or a 
\textbf{time synchronized} combination 
of audio and video. \textbf{Our goal:} 
estimate identifiable physical parameters 
$\boldsymbol{\theta}\in\mathbb{R}^{K}$ 
governing system dynamics.

\noindent A \textbf{Dynamical system} is 
represented as a set of $D$ dimensional 
states $\mathbf{x}(t)\in\mathbb{R}^{D}$ 
evolves via parametric ODEs:

\begin{equation}
\frac{d\mathbf{x}(t)}{dt}
= f\!\big(\mathbf{x}(t),\,
\mathbf{u}(t);\boldsymbol{\theta}\big),
\label{eq:phys_ode}
\end{equation}
where $\mathbf{u}(t)$ are exogenous inputs 
often not available (occluded) in visual 
modalities, and $f$ encodes 
domain-specific physics.

\noindent{\bf Invariants} are 
time-independent reference quantities 
$\psi$, such as coordinate-frame origins, 
camera-to-world alignment, and initial 
states, that parameterize the transformation 
$X_{\mathrm{world}} = 
g(X_{\mathrm{obs}};\psi)$. 
Although they do not evolve with the dynamics, 
they must be jointly estimated to express 
the recovered model in a physically 
meaningful frame.

\noindent\textbf{Implicit dynamics.} We assume 
that only a subset of state variables 
$\mathbf{x}$ can be measured denoted by a 
measurement matrix $M$ which is a 
$D\times D$ diagonal matrix with 
$M(i,i) = 1 (0)$ if the $i^{th}$ state 
variable is measured (or not).

\begin{figure*}[t]
\centering
\includegraphics[width=\textwidth]
{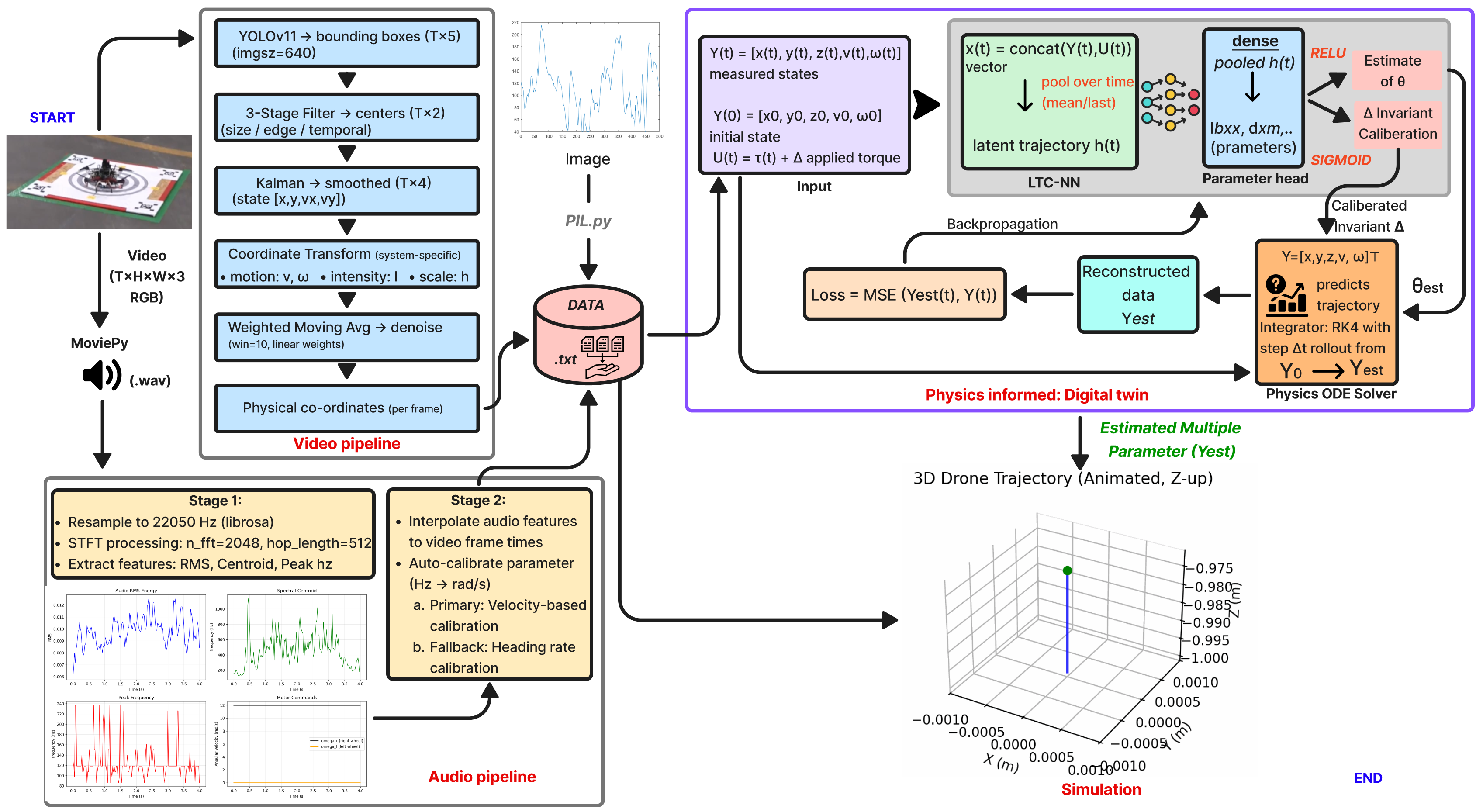}
\caption{\textbf{EMMA Architecture.} 
Multi-modal inputs (video, audio and image) 
are processed through specialized pipelines 
into unified temporal representations. 
An LTC-NN learns latent dynamics $h(t)$ 
and predicts physical parameters $\theta$. 
A differentiable physics simulator validates 
predictions, enabling end-to-end gradient 
flow. The framework successfully reconstructs 
digital twin drone (example shown) from 
multi-modal observations.}
\label{fig:teaser2}
\end{figure*}

\noindent\textbf{Parameter estimation.} 
Given an estimated $\boldsymbol{\theta}_{est}$ 
and $\mathbf{x}_0$, we simulate 
Eq.~\ref{eq:phys_ode} to obtain trajectory 
$\mathbf{x}^{\text{sim}}(t)$ and observables 
$\mathbf{y}^{\text{sim}}_t=
P_{\text{obs}}\!\big(
\mathbf{x}^{\text{sim}}(t)\big)$. 
We minimize:
\begin{equation}
\min_{\boldsymbol{\theta}\,
(\!,\,\mathbf{x}_0)}
\;\frac{1}{T}\sum_{t=1}^T
\big\|\mathbf{y}_t
-\mathbf{y}^{\text{sim}}_t\big\|_2^2
\;+\;\mathcal{R}(\boldsymbol{\theta}_{est}),
\label{eq:objective}
\end{equation}
where $\mathcal{R}$ enforces physical 
validity (positivity, bounds).

\subsection{Architecture Overview}
\label{sec:architecture}

EMMA consists of three stages 
(Figure~\ref{fig:teaser2}): 
\textit{(i)} \textbf{unified multi-modal 
feature extraction} from video, audio, 
and images producing time-aligned sequences 
$\{\mathbf{x}(t)\}_{t=1}^T$ through 
modality-specific processing with temporal 
synchronization; 
\textit{(ii)} \textbf{Liquid Time-Constant 
network} modeling continuous-time latent 
dynamics with input-dependent time constants, 
outputting hidden trajectories 
$\mathbf{h}(t) \in \mathbb{R}^{H}$, 
where $H$ is the hidden state dimension; 
\textit{(iii)} \textbf{multi-parameter 
estimation} via sequence-to-sequence 
prediction with temporal averaging, 
yielding 
$\bar{\boldsymbol{\theta}} \in \mathbb{R}^K$, 
resulting in a digital twin validated 
through physics simulation.

\subsection{Unified Multi-modal Feature 
Extraction}
\label{sec:multimodal}

Our unified framework processes heterogeneous 
modalities through specialized pipelines that 
produce temporally-aligned feature sequences.

\noindent\textbf{Video pipeline.} From frames 
$\{I_t\}_{t=1}^T$, we extract object 
trajectories through five stages. 
\textit{Stage 1 (Detection)}: 
YOLOv11~\cite{jocher2024yolo} detects 
bounding boxes with confidence threshold 0.85 (higher thresholds cause significant frame loss).
\textit{Stage 2 (Filtering)}: Three-stage 
filtering computes centers, removes edge 
detections (10px threshold), enforces 
temporal stability. 
\textit{Stage 3 (Smoothing)}: 
Kalman filter~\cite{kalman1960filter} with 
state $[x, y, v_x, v_y]$ reduces jitter. 
\textit{Stage 4 (Transform)}: For pendulum, 
pixel-to-angular conversion 
$\theta = \arctan\left(
\frac{y - y_p}{x - x_p}\right)$; 
for motion, calibrated transforms. 
\textit{Stage 5 (Denoising)}: Weighted moving 
average (window 10) smooths coordinates. 
Output: physical coordinates 
$\mathbf{p}(t) \in \mathbb{R}^{d_v}$.
We also evaluate an unsupervised 
Farneback optical flow alternative 
that achieves comparable accuracy 
without any pretrained detector, 
confirming that EMMA's core contribution 
lies in the LTC physics layer rather 
than the specific feature extractor 
(see supplementary Table~S4).

\textbf{Audio pipeline} has the 
following components:

\textit{a) Signal processing:} Raw audio 
signals $\{A_t\}_{t=1}^T$ are recorded at 
44.1\,kHz and resampled to
22.05\,kHz. We compute the STFT (FFT size 
2048, hop 512) using
\texttt{librosa}~\cite{mcfee2015librosa}, 
extracting RMS energy, spectral
centroid, and dominant spectral peak frequency. 
All features are temporally
aligned to video timestamps via an 
auto-calibration module, yielding acoustic
feature vectors 
$\mathbf{w}(t) \in \mathbb{R}^{d_a}$.

\textit{b) Datasheet-based linearity prior.} 
For non-flying rotors and rover wheels, 
manufacturer datasheets provide the nominal 
tonal frequency at a given RPM as well as 
the incremental change in frequency per unit 
increase in rotational speed. Empirically, 
the dominant tonal component of the acoustic 
signal varies approximately linearly with 
rotor/wheel speed in the non-flight regime. 
We therefore impose a linear audio-speed 
prior $f_{\mathrm{tone}}(t) \approx 
\alpha\, v(t) + \beta$, where 
$f_{\mathrm{tone}}(t)$ is the extracted 
spectral peak frequency obtained from 
$\mathbf{w}(t)$ and $v(t)$ is the physical 
rotational speed. The affine transform 
parameters $\alpha$ and $\beta$ are 
calibration parameters (invariants) that 
are learned through the LTC-NN based 
component of EMMA.

\textbf{Image modality.} Images 
$\{M_t\}_{t=1}^T$ from sensors, thermal 
cameras, or measurement devices undergo 
feature extraction through CNNs or direct 
processing pipelines. For thermal imaging, 
we extract temperature distributions and 
gradients. For sensor images (e.g., medical 
scans, microscopy), we extract intensity 
profiles and spatial patterns. Output: 
image feature vectors 
$\mathbf{m}(t) \in \mathbb{R}^{d_m}$. 
We apply lightweight preprocessing with 
Pillow (PIL): load PNG or JPEG files, 
crop regions of interest, perform mode 
conversion (RGB to L and back), and 
normalize contrast before converting 
arrays to tensors. For chart images, 
we combine PIL pixel access with OpenCV 
masks to isolate the curve color and 
discretize it into $(x,y)$ time series 
points for downstream modules.

\textbf{Unified temporal alignment.} 
All modality features undergo temporal 
interpolation to align with video frame 
timestamps, creating a unified temporal 
grid. Features concatenate to form 
multi-modal state vectors:
\begin{equation}
\mathbf{x}(t) = [\mathbf{p}(t); 
\mathbf{w}(t); \mathbf{m}(t)] 
\in \mathbb{R}^{D_{\text{in}}},
\label{eq:multimodal_concat}
\end{equation}
where $D_{\text{in}} = d_v + d_a + d_m$ 
depends on available modalities. Missing 
modalities are handled through zero-padding 
or learned embeddings.

\textbf{Spatial encoding for multimodal 
fusion.} For systems with multiple scenarios, 
temporal trajectories discretized into 
$N_{\text{spatial}}=100$ samples per modality. 
This spatial encoding enables efficient 
processing of dense trajectory information 
while maintaining temporal coherence across 
heterogeneous inputs, producing unified 
representations 
$\mathbf{x}_{\text{in}}(t) 
\in \mathbb{R}^{100}$.

\subsection{LTC Network for 
Continuous-Time Dynamics}
\label{sec:ltc}

\begin{figure}[t]
\centering
\includegraphics[width=\linewidth]
{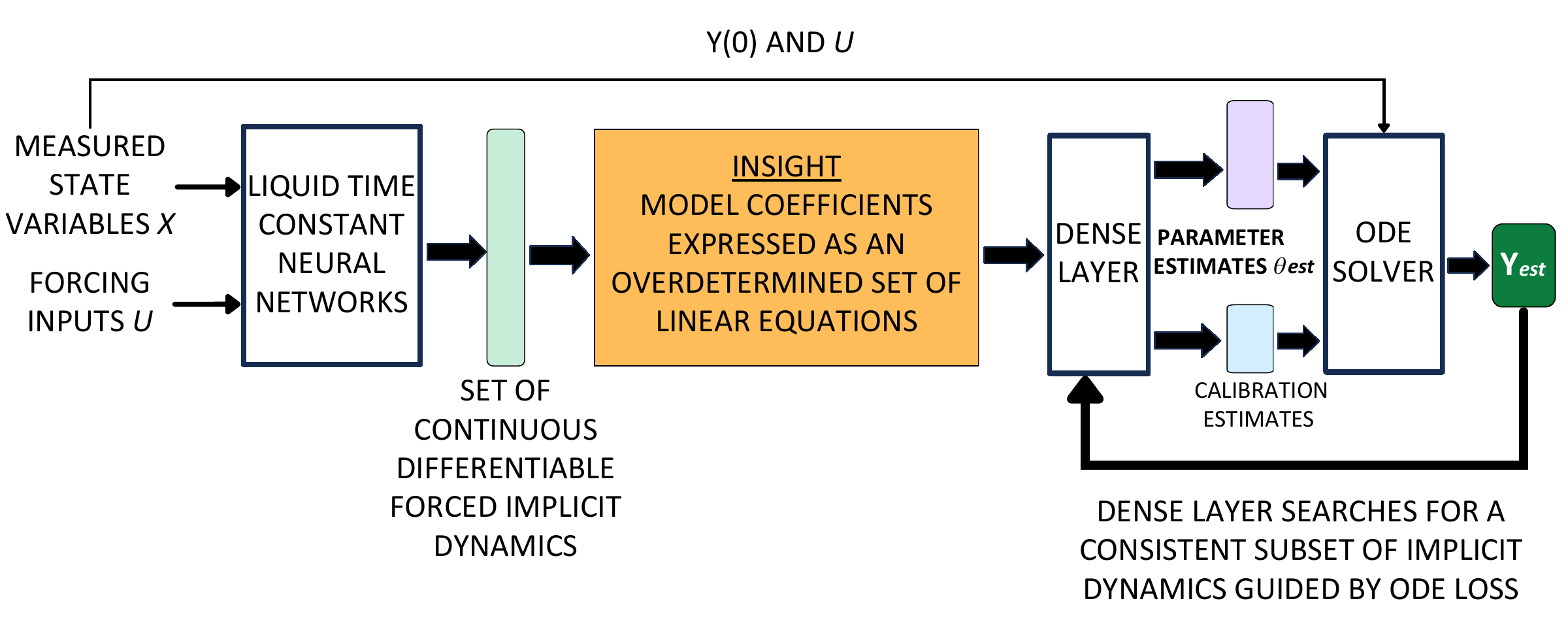}
\caption{Intuition for using LTC-NN 
for parameter estimation.}
\label{fig:LTC}
\end{figure}

The parameter estimation component of EMMA 
uses two fully connected layers 
(Figure \ref{fig:LTC}): LTC-NN, which 
solves two problems of implicit dynamics 
and forcing inputs, and a dense layer that 
learns the model parameter estimates as 
nonlinear function of the LTC-NN hidden 
outputs and simultaneously calibrates the 
invariant estimates.

\noindent{\bf How does LTC-NN tackle 
forcing inputs?} We use ncps 
library~\cite{hasani2021ncps} with 64 
hidden units. Each LTC-NN cell $i$ 
implements:
\begin{equation}
    \label{eqn:Str}
    \scriptsize
    \frac{dh_i}{dt} = 
    \underbrace{\frac{-h_i}
    {\frac{\tau_i}{1+
    \tau_i f_{NN}(h_i,u,t,w_{NN})}}}
    _{\text{models forcing inputs}} 
    + \underbrace{f_{NN}(h_i,u,t,w_{NN})A}
    _{\text{models physics consistent 
    dynamics}},
\end{equation}
where $h_i(t) \in \mathbf{h}(t) 
\in \mathbb{R}^{64}$ is hidden state 
encoding latent dynamics across all 
modalities, $\tau_i(t) > 0$ is a time 
constant~\cite{hasani2021ncps}
enabling adaptive response to multi-modal 
inputs, $w_{NN}$ and $A$ are recurrent 
weights, $u(t)$ are inputs, and 
$f_{NN}(,.,.,.,.)$ is the forward pass 
typically perceptron. 
Equation \ref{eqn:Str} shows that the 
forward pass of LTC-NN inherently has an 
input-dependent time constant that helps 
in modeling forcing inputs.
We compare LTC against Neural 
ODE~\cite{chen2018neuralode} and 
CT-GRU~\cite{debrouwer2019ctgru} in the 
supplementary (Table~S7); under forcing 
inputs, LTC outperforms Neural ODE by 
25\% and CT-GRU by 5\% in average 
parameter error, validating the importance 
of input-dependent time constants.

\noindent{\bf How does LTC-NN model implicit 
dynamics?} The second part of 
Equation \ref{eqn:Str} is a set of hidden 
outputs that are themselves expressed as a 
consistent set of differential equations. 
These are much more variables than the 
required number of states in the system and 
hence carry the capability to model multiple 
implicit dynamics.

\noindent{\bf How does EMMA obtain parameters?}
The dense head in Fig.~\ref{fig:LTC} maps 
measured and latent dynamics to physical 
parameters via a nonlinear readout with 
\textsc{sigmoid} activations, leveraging the 
universal approximation property of 
feedforward 
networks~\cite{cybenko1989approximation,
hornik1989multilayer}.
This readout can be interpreted as a 
data-driven modal decomposition of latent 
trajectories, in the spirit of Dynamic Mode 
Decomposition and its Koopman extensions 
(EDMD) and variants with control 
(DMDc)~\cite{schmid2010dmd,
williams2015edmd,
proctor2016dmdc,
kutz2016dmd}.
The predicted parameters are injected into 
the known ODE and the whole system is trained 
end-to-end via differentiable simulation, 
aligning with physics-informed and 
system-identification 
methodologies~\cite{raissi2019physics,
brunton2016discovering,
kaiser2018sindyc}.
Because the \textsc{sigmoid} range is 
$(0,1)$, we apply the denormalization in 
Eq.~\eqref{eq:denorm}.

\noindent{\bf How does EMMA obtain calibration 
parameters?} For calibration, we add more 
cells in the dense layer than the number of 
dynamical parameters. These additional cells 
have RELU activation function and change 
linearly with hidden inputs and loss gradient. 
These hidden outputs are used for modeling 
the calibration parameters.

\textbf{Denormalization.} We map the dense 
layer outputs to physical scales via:
\begin{equation}
\scriptsize
    \theta_k = \left(1 + \left(
    0.5 - \bar{\boldsymbol{\theta}}_k
    \right) \cdot \frac{95}{100}\right) 
    \cdot \theta_k^{\text{nom}}.
    \label{eq:denorm}
\end{equation}

\subsection{Loss Functions and Training}
\label{sec:training}


\textbf{Physics-informed loss.} Total loss 
combines trajectory accuracy, calibration 
parameter along with constraints:
\begin{equation}
\scriptsize
    \mathcal{L}_{\text{total}} = 
    \mathcal{L}^{cal}_{\text{traj}} + 
    \lambda_{\text{param}} 
    \mathcal{L}_{\text{param}}.
    \label{eq:loss_total}
\end{equation}
EMMA is \textit{unsupervised} with respect to 
parameters: no ground-truth parameter 
values are used during training. Learning 
is driven solely by the physics-based loss 
(Eq.~\ref{eq:loss_total}).

\textbf{Calibrated trajectory loss.} 
Measures discrepancy across measured 
variables:

\begin{scriptsize}
\begin{equation}
\mathcal{L}^{cal}_{\text{traj}}
= \sum_{i=1}^{n}M_{ii}
\frac{1}{T_{\text{sim}}} 
\sum_{t=1}^{T_{\text{sim}}}
  \lVert x_{i}(t)- \gamma_i 
  - x_{i,\text{sim}}(t) \rVert^2,
\label{eq:loss_traj}
\end{equation}
\end{scriptsize}
where $\gamma_i$ is given by a dense layer 
RELU output if the $i^{th}$ state variable 
requires calibration else $\gamma_i = 0$.

\textbf{Parameter constraints.}
ReLU penalties enforce valid parameter ranges:

\begin{scriptsize}
\begin{equation}
\begin{split}
&\mathcal{L}_{\text{param}}= 
\sum_{i=1}^{i=K}\, w_p(i)\,
\text{ReLU}(-\theta_i) \\
&+ w_l(i)\,\text{ReLU}(\theta_i 
- l_i) 
+ w_{up}(i)\,
\text{ReLU}(\theta_i - up_i),
\end{split}
\label{eq:loss_param}
\end{equation}
\end{scriptsize}

where $w_p$, $w_l$, and $w_{up}$ are 
penalties for violating positivity, lower 
bound and upper bound respectively.

\textbf{Optimization.}
AdamW~\cite{loshchilov2019adamw} with cosine 
annealing~\cite{loshchilov2017sgdr}, all 
parameters listed in 
Table~\ref{tab:hyperparameters}.

\textbf{Implementation.} We implement EMMA 
in PyTorch~\cite{paszke2019pytorch} using the
\texttt{ncps} library~\cite{hasani2021ncps} 
for LTC networks with input size 100, 
64 hidden units,
sequence-to-sequence output, and 6 ODE 
unfolding steps. Video processing uses 
YOLOv11~\cite{jocher2024yolo}
for detection and 
OpenCV~\cite{bradski2000opencv} for tracking. 
Audio features are extracted with
librosa~\cite{mcfee2015librosa} and 
MoviePy~\cite{moviepy}.


\section{Evaluation}
\label{sec:experiments}

\begin{table*}[t]
\centering
\scriptsize
\setlength{\tabcolsep}{3.5pt}

\begin{tabular}{@{}l
                *{3}{c} G *{3}{c} G 
                *{3}{c} G *{3}{c} G 
                *{3}{c}@{}}
\toprule
\rowcolor{gray!15}
& \multicolumn{3}{c}{\textbf{Pendulum}} 
& \multicolumn{3}{c}{\textbf{Torricelli}} 
& \multicolumn{3}{c}{\textbf{Sliding Block}} 
& \multicolumn{3}{c}{\textbf{LED}} 
& \multicolumn{3}{c@{}}{\textbf{Free Fall}} \\
\rowcolor{gray!8}\scriptsize
& \multicolumn{3}{c}{$L$ [m]} 
& \multicolumn{3}{c}{$k$ [$\sqrt{m}/s$]} 
& \multicolumn{3}{c}{$a$ [$m/s^2$]} 
& \multicolumn{3}{c}{$\gamma$} 
& \multicolumn{3}{c@{}}{$a$ [$m/s^2$]} \\
\cmidrule(lr){2-4}\cmidrule(lr){5-7}
\cmidrule(lr){8-10}\cmidrule(lr){11-13}
\cmidrule(lr){14-16}
\textbf{Method} 
& 45cm & 90cm & 150cm 
& Small & Med & Large 
& Low & Med & High 
& Low & Med & High 
& Small & Med & Large \\
\midrule
PAIG~\cite{jaques2020physics}
& 1.01 & 1.01 & 1.01 
& 0.99 & 0.99 & 0.97 
& 0.35 & 0.38 & 0.37 
& --- & --- & --- 
& --- & --- & --- \\
\sdrow
& $\pm$0.03 & $\pm$0.04 & $\pm$0.04 
& $\pm$0.01 & $\pm$0.01 & $\pm$0.02 
& $\pm$0.03 & $\pm$0.02 & $\pm$0.04 
& & & & & & \\

NIRPI~\cite{hofherr2023neural}
& 0.77 & 0.84 & 0.63 
& 0.21 & 0.14 & 0.16 
& $-0.09$ & $-0.01$ & $-0.06$ 
& --- & --- & --- 
& --- & --- & --- \\
\sdrow
& $\pm$0.33 & $\pm$0.53 & $\pm$0.38 
& $\pm$0.03 & $\pm$0.04 & $\pm$0.01 
& $\pm$0.88 & $\pm$0.5 & $\pm$0.01 
& & & & & & \\

Delfys~\cite{garcia2025lfv}
& 0.51 & 1.07 & 1.30 
& \textbf{0.0094} & \textbf{0.0132} & 0.0167 
& 1.29 & 2.70 & 3.44 
& 2.24 & 0.97 & 0.41 
& 15.0 & 9.51 & \textbf{10.22} \\
\sdrow
& $\pm$0.01 & $\pm$0.2 & $\pm$0.02 
& \textbf{$\pm$4e$^{-4}$} 
& \textbf{$\pm$5e$^{-4}$} 
& $\pm$4e$^{-4}$ 
& $\pm$0.1 & $\pm$0.09 & $\pm$0.19 
& $\pm$0.36 & $\pm$0.04 & $\pm$0.04 
& $\pm$2.1 & $\pm$1.27 
& \textbf{$\pm$1.21} \\

PySINDy~\cite{brunton2016discovering}
& 0.28 & 0.77 & 1.24 
& 0.035 & 0.027 & 0.068 
& 2.13 & 2.64 & 2.63 
& 1.74 & 0.73 & 0.37 
& 2.64 & 6.66 & 1.50 \\
\sdrow
& $\pm$0.18 & $\pm$0.52 & $\pm$1.3 
& $\pm$0.029 & $\pm$0.011 & $\pm$0.035 
& $\pm$0.71 & $\pm$0.46 & $\pm$0.47 
& $\pm$0.62 & $\pm$0.0 & $\pm$0.08 
& $\pm$1.82 & $\pm$0.0 & $\pm$1.12 \\

\midrule
\textbf{EMMA}
& \textbf{0.50} & \textbf{0.86} 
& \textbf{1.50} 
& 0.0093 & \textbf{0.0132} 
& \textbf{0.0163} 
& \textbf{1.41} & \textbf{2.27} 
& \textbf{3.14} 
& \textbf{2.29} & \textbf{0.91} 
& \textbf{0.45} 
& \textbf{5.2} & \textbf{9.95} & 10.29 \\
\sdrow
& \textbf{$\pm$0.039} 
& \textbf{$\pm$0.073} 
& \textbf{$\pm$0.004} 
& $\pm$0.0004 
& \textbf{$\pm$0.0008} 
& \textbf{$\pm$0.0009} 
& \textbf{$\pm$0.018} 
& \textbf{$\pm$0.029} 
& \textbf{$\pm$0.050} 
& \textbf{$\pm$0.0} 
& \textbf{$\pm$0.0} 
& \textbf{$\pm$0.0} 
& \textbf{$\pm$1e$^{-5}$} 
& \textbf{$\pm$0.0} & $\pm$0.69 \\

\midrule
GT & 0.45 & 0.90 & 1.50 
& 0.0095 & 0.0128 & 0.0162 
& 1.441 & 2.300 & 3.141 
& 2.3 & 0.92 & 0.46 
& 9.8 & 9.8 & 9.8 \\
\bottomrule
\end{tabular}
\caption{Comprehensive baseline comparison 
across multiple physical systems. EMMA 
demonstrates competitive or superior 
parameter estimation accuracy across 
motion-based dynamics (Pendulum, Torricelli, 
Sliding Block) and extends to non-motion 
domains (LED decay, Free Fall). 
Mean $\pm$ standard deviation over five 
videos per configuration. 
Best results are in bold.}
\label{tab:comprehensive_baseline}

\end{table*}

We have three evaluation objectives:

\noindent{\bf Experiment A: Multi parameter 
study}: we test whether EMMA can accurately 
extract multiple parameters from video. 
\underline{Benchmarks:} we use the 
Delfys~\cite{garcia2025lfv} dataset for this 
experiment. We evaluate on five canonical 
physical systems with known ground-truth 
parameters. \textbf{Pendulum} dynamics 
estimate length $L$ and damping coefficient 
$\tau$ across three configurations 
(45cm, 90cm, 150cm) with 5 videos each. 
\textbf{Torricelli drainage} estimates 
the drainage constant $k$ for three 
container sizes. \textbf{Sliding block} 
estimates acceleration and friction across 
three incline slopes. \textbf{LED decay} 
estimates decay rate $\gamma$ across three 
temporal regimes. \textbf{Free fall} 
estimates gravitational acceleration $g$ 
across three object sizes. Each configuration 
includes 5 videos, totaling 75 benchmark 
videos. Detailed information about the 
dynamical equations are provided in the 
supplementary documents. 
\underline{Baselines:} We compare to 
video-based parameter inference methods 
PAIG~\cite{jaques2020physics}, 
NIRPI~\cite{hofherr2023neural}, 
Delfys~\cite{garcia2025lfv} for comparison 
with single parameter extraction and 
PySINDy~\cite{brunton2016discovering} for 
multi-parameter extraction. 
\underline{Metrics:} Since these examples 
have at most two parameters, we compare 
the raw values of the extracted parameters 
with standard deviation.

\noindent{\bf b) Parameter extraction under 
implicit and forced dynamics}, we test 
whether EMMA can accurately extract 
parameters from multi-modal audio-visual 
data with implicit dynamics and forcing 
inputs for real world videos. 
\underline{Benchmarks:} For real-world 
validation, we test on a differential-drive 
RC rover. It has 9 parameters including 
wheel radius, baseline width, and motor 
constants, while only 5 have known ground 
truths obtained from datasheets. We also 
utilize a 6-DoF quadcopter with 12 parameters 
out of which 7 are known. The rover video 
is collected in our lab, while the quadcopter 
video is collected from the lab Youtube page 
of University of Pennsylvania with ground 
truth obtained from datasheets. Detailed 
information about ground truth information 
and datasheet are provided in supplement. 
\underline{Baselines:} Since this is the 
first work to perform parameter extraction 
from multi-modal audio video data under 
forced inputs and implicit dynamics, there 
are no baselines for this task. We compare 
the extracted parameters with ground truth. 
\underline{Metrics:} Raw values are compared 
with ground truth.

\noindent{\bf c) Parameter extraction from 
charts of simulations}, we test whether 
EMMA can extract multiple known parameters 
from simulation chart images. These 
experiments demonstrate that EMMA can 
extract models from at least three different 
modalities. \underline{Benchmarks:} We use 
the figures generated from the simulator 
available for each of the case studies 
reported 
in~\cite{kaiser2018sindyc,
desilva2020pysindy}. 
There are four case studies of F8 Crusader, 
Lotka Volterra, Lorenz oscillator and 
HIV therapy. The exact image generation 
mechanism and the detailed dynamics is 
discussed in supplement. In addition we use 
a medical case study of automated insulin 
delivery system for Type 1 Diabetes. The 
data for this purpose was generated using 
the UVA/PADOVA Type 1 Diabetes 
Simulator~\cite{dallaman2007meal}. 
The underlying dynamics is governed by 
Bergman Minimal Model with only sensed 
variable is glucose and all other variables 
are implicit. \underline{Baselines:} PySindy 
is the main comparator in this case. However, 
since PySindy does not have a image pipeline, 
we use the EMMA image processing pipeline 
for both the evaluations of EMMA and PySindy 
ensuring fair comparison. 
\underline{Metrics:} Since the number of 
parameters extracted are many and different 
for each application, for consideration of 
space we only report the root mean square 
error (RMSE) in reconstruction $x_{rmse}$ 
of the data from learned parameters and the 
RMSE in parameter estimation $\theta_{rmse}$.

\noindent\textbf{Inputs and preprocessing.} 
EMMA accepts any subset of video, audio, 
and sensor modalities. Video processing 
employs YOLOv11 for object detection 
(confidence threshold 0.85, image size 
640 pixels), followed by Kalman filtering 
with state vector $[x, y, v_x, v_y]$ and 
pixel-to-metric calibration via 
domain-specific coordinate transformations. 
Audio extraction uses MoviePy at 44,100 Hz 
with librosa spectral features (RMS energy, 
spectral centroid, peak frequency). All 
inputs undergo z-score normalization using 
training statistics; parameters are 
denormalized at inference via 
Eq.~\eqref{eq:denorm}.

\begin{table}[t]
\centering
\scriptsize
\begin{tabular}{@{}lc@{}}
\toprule
\textbf{Parameter} & \textbf{Value} \\
\midrule
\textit{LTC Network} & \\
Units, ODE unfolds, $\epsilon$ 
  & 64, 6, $10^{-8}$ \\
\midrule
\textit{AdamW Optimizer} & \\
$\eta$, $\lambda_{wd}$, $\epsilon$, 
$\beta_1$, $\beta_2$ 
  & $5\times10^{-3}$, $10^{-4}$, 
  $10^{-8}$, 0.9, 0.999 \\
\midrule
\textit{Cosine Schedule} & \\
$T_0$, $T_{mult}$, $\eta_{min}$ 
  & 10, 2, $10^{-6}$ \\
\midrule
\textit{Training} & \\
Window ($T$), Stride, Batch, 
Patience, $p_{drop}$ 
  & 16, 1, 32, 40 epochs, 0.3  \\
\bottomrule
\end{tabular}
\caption{Training hyperparameters 
and configuration.}
\label{tab:hyperparameters}
\end{table}

\noindent\textbf{Reporting and splits.} 
For each benchmark configuration, we 
compute mean $\pm$ standard deviation 
over 5 videos. When sensors or audio are 
absent, we report results using available 
modalities. We fix random seeds and 
document all hyperparameters for 
reproducibility as detailed in 
Table \ref{tab:hyperparameters}.
Multi-seed statistical validation 
(5 seeds) is provided in 
supplementary Tables~S5 and~S6 
for rover and drone respectively.

\begin{table*}[t]
\centering
\begingroup
\scriptsize
\setlength{\tabcolsep}{3.2pt}
\renewcommand{\arraystretch}{1.12}
\setlength{\abovetopsep}{3pt}
\setlength{\belowrulesep}{1.2pt}
\setlength{\aboverulesep}{0.8pt}
\setlength{\belowbottomsep}{3pt}

\begin{minipage}[t]{0.495\textwidth}\centering
\textbf{(a) Pendulum}\par\vspace{2pt}
\resizebox{\linewidth}{!}{%
\begin{tabular}{@{}lcccccc@{}}
\toprule
\multirow{2}{*}{\textbf{Length}} &
\multicolumn{2}{c}{\textbf{EMMA}} &
\multicolumn{2}{c}{\textbf{PySINDy}} &
\multicolumn{2}{c}{\textbf{GT}}\\
\cmidrule(lr){2-3}\cmidrule(lr){4-5}
\cmidrule(l){6-7}
 & $L$ (m) & $\tau$ (1/s) 
 & $L$ (m) & $\tau$ (1/s) 
 & $L$ (m) & $\tau$ (1/s)\\
\midrule
45cm  & $0.507 \pm 0.039$ 
     & $0.055 \pm 0.026$ 
     & $0.36 \pm 0.06$  
     & $0.73 \pm 0.57$ 
     & 0.45 & 0.05 \\
90cm  & $0.859 \pm 0.073$ 
     & $0.045 \pm 0.011$ 
     & $0.81 \pm 0.04$  
     & $0.01 \pm 0.03$ 
     & 0.9 & 0.05 \\
150cm & $1.501 \pm 0.004$ 
     & $0.050 \pm 0.001$ 
     & $1.51 \pm 0.11$ 
     & $0.00 \pm 0.00$ 
     & 1.5 & 0.05 \\
\bottomrule
\end{tabular}%
}
\end{minipage}\hfill
\begin{minipage}[t]{0.495\textwidth}\centering
\textbf{(b) Sliding Block}\par\vspace{2pt}
\resizebox{\linewidth}{!}{%
\begin{tabular}{@{}lcccccc@{}}
\toprule
\multirow{2}{*}{\textbf{Slope}} &
\multicolumn{2}{c}{\textbf{EMMA}} &
\multicolumn{2}{c}{\textbf{PySINDy}} &
\multicolumn{2}{c}{\textbf{GT}}\\
\cmidrule(lr){2-3}\cmidrule(lr){4-5}
\cmidrule(l){6-7}
 & $\alpha$ ($^\circ$) & $\mu$ 
 & $\alpha$ ($^\circ$) & $\mu$ 
 & $\alpha$ ($^\circ$) & $\mu$ \\
\midrule
Low    & $19.92 \pm 0.20$ 
       & $0.208 \pm 0.003$ 
       & $24.00 \pm 4.18$ 
       & $0.21 \pm 0.00$ 
       & 20.0 & 0.20 \\
Medium & $24.72 \pm 0.38$ 
       & $0.205 \pm 0.004$ 
       & $27.00 \pm 2.74$ 
       & $0.21 \pm 0.00$ 
       & 25.0 & 0.20 \\
High   & $29.81 \pm 0.42$ 
       & $0.204 \pm 0.004$ 
       & $27.00 \pm 2.74$ 
       & $0.21 \pm 0.00$ 
       & 30.0 & 0.20 \\
\bottomrule
\end{tabular}%
}
\end{minipage}

\vspace{8pt}

\begin{minipage}[t]{0.495\textwidth}\centering
\textbf{(c) Rover (5 parameters)}\par\vspace{2pt}
\begin{tabular}{@{}lccp{2.7cm}@{}}
\toprule
\textbf{Parameter} & \textbf{Ground Truth} 
& \textbf{Value} & \textbf{Description} \\
\midrule
$a$   & $0.178$ m  & $0.196$ m  
      & X-arm length   \\
$b$   & $0.144$ m  & $0.134$ m  
      & Y-arm length   \\
$r$   & $0.201$ m  & $0.223$ m  
      & Wheel radius   \\
$m$   & $26.88$ kg & $24.44$ kg 
      & Mass           \\
$CM$  & $0.112$ m  & $0.120$ m  
      & CoM height     \\
\bottomrule
\end{tabular}
\end{minipage}\hfill
\begin{minipage}[t]{0.495\textwidth}\centering
\textbf{(d) Drone (7 parameters)}%
\par\vspace{2pt}
\begin{tabular}{@{}lccp{2.7cm}@{}}
\toprule
\textbf{Parameter} & \textbf{Ground Truth} 
& \textbf{Value} & \textbf{Description} \\
\midrule
$k_{Th}$  & 1.1   & 1.017 
          & Thrust coef.      \\
$k_{To}$  & 1.3   & 1.501 
          & Torque coef.      \\
$k_p$     & 0.91  & 1.007 
          & Motor gain        \\
$\tau_2$  & 0.012 & 0.015 
          & Motor time const. \\
$d_{xm}$  & 0.18  & 0.158 
          & X-arm length      \\
$d_{ym}$  & 0.20  & 0.173 
          & Y-arm length      \\
$d_{zm}$  & 0.07  & 0.051 
          & Z-arm offset      \\
\bottomrule
\end{tabular}
\end{minipage}
\endgroup

\caption{(4.a, 4.b) Side-by-side comparison of EMMA 
and PySINDy parameter estimates across 
shared physical systems. Values are 
mean $\pm$ standard deviation over 
five videos. (4.c, 4.d) Rover and drone parameters 
estimated by EMMA, compared against 
ground truth (GT).}
\label{tab:emma-pysindy-gt}
\end{table*}

\subsection{Benchmark Results}

\noindent\textbf{Experiment A: Single + 
Multi-parameter study on Video only:} 
Table~\ref{tab:comprehensive_baseline} 
summarizes results across the five benchmark 
systems for single parameter estimation. 
EMMA achieves competitive or superior 
parameter accuracy. Relative to video 
baselines and PySINDy, EMMA consistently 
reduces parameter error.

\noindent\textit{Pendulum:} EMMA recovers 
length $L$ and damping $\tau$ close to 
ground truth across 45/90/150 cm 
configurations. Video baselines show larger 
bias in $L$ at extreme lengths. PySINDy's 
derivative estimation is sensitive to noise 
and occlusion, producing parameter estimates 
with high variance.

\noindent\textit{Torricelli:} EMMA accurately 
estimates the drainage constant $k$ despite 
the $\sqrt{h}$ nonlinearity. PySINDY method 
struggles to represent fractional powers, 
leading to systematic errors. Our 
physics-constrained loss stabilizes learning 
and tightly matches ground truth across 
container sizes with low variance 
($\pm 0.0004$ to $\pm 0.0009$).

\noindent\textit{Sliding block:} EMMA improves 
estimation of acceleration and friction across 
low, medium, and high slopes, producing lower 
error than video baselines and PySINDy. 
Parameter estimates remain stable across 
different slope configurations.

\noindent\textit{LED decay:} EMMA closely 
matches decay rates across fast, medium, and 
slow regimes with low variance across videos. 
Error remains stable under moderate measurement 
noise, demonstrating robustness to realistic 
lighting variations and camera auto-adjustment 
artifacts.

\noindent\textit{Free fall:} EMMA recovers 
gravitational acceleration $g$ across object 
sizes and operating conditions. PySINDY 
baseline is sensitive to discrete 
differentiation and frame-rate variations, 
resulting in larger errors. EMMA's 
continuous-time formulation via LTC networks 
handles irregular sampling naturally.

\noindent\textit{Multi-parameter Settings:} 
Table \ref{tab:emma-pysindy-gt} (a and b) 
shows the performance of EMMA in extracting 
multiple parameters from the benchmark 
examples available in the Delfys dataset. 
The only comparator here is PySINDY and EMMA 
is superior in estimating parameters.

\noindent\textit{Takeaways:} Across five 
benchmarks spanning diverse physical regimes, 
EMMA's physics-informed training and 
continuous-time dynamics deliver accurate 
parameter estimates. The framework outperforms 
baseline methods in most settings while 
eliminating mask requirements and pixel-space 
reconstruction overhead. Quantitative results 
with mean $\pm$ std by configuration are 
shown in 
Table~\ref{tab:comprehensive_baseline}. 
EMMA could learn invariant parameters such 
as hanging point of the pendulum or pendulum 
starting point and did not use these as 
prior information.
EMMA also converges accurately under 
expanded initialization bounds 
(200\% range) in 5 of 6 configurations, 
confirming robustness to poor initialization 
(supplementary Table~S8).

\noindent{\bf Experiment B: Multi parameter 
extraction under implicit and forced 
dynamics}: 
Table \ref{tab:emma-pysindy-gt} (c and d) 
shows the dynamical parameters estimated by 
EMMA and compares with the ground truth 
parameters in the two real world case studies 
of drone and rover. We could only compare 
EMMA estimated parameters with the available 
ground truths. EMMA has an average error of 
15.9\% $\pm$ 7.4\% in estimating 
all measurable parameters of the 
drone, while it has 
8.8\% $\pm$ 1.7\% for the rover 
example. For the Rover the center of mass 
(CoM) height and the wheel radius are 
parameters related to implicit dynamics, 
while for the drone example the thrust 
coefficient, torque coefficient, motor gain 
and motor time constant are implicit dynamics 
parameters. EMMA's performance remained 
stable across parameters related to implicit 
or measured parameters.

\noindent\textit{Takeaways:} Even under 
forcing inputs from the rover or drone 
controller, EMMA could extract parameters related to both measured and implicit 
dynamics. EMMA also performed good calibration 
since it did not use idle wheel power or 
quadrotor idle rotational speed as inputs 
or coordinate space origin as input 
instead EMMA learned the most appropriate 
invariants.
Audio noise robustness experiments 
(supplementary Table~S9) confirm that 
injecting Gaussian noise at SNR levels 
down to 5\,dB causes less than 1.1\% 
variation in all estimated rover parameters.

\noindent{\bf Experiment C: Parameter 
extraction from simulation charts:} In this 
experiment for the simulation charts we 
executed EMMA and PySINDY under two 
conditions: a) implicit dynamics, where the 
image pipeline was applied on only the chart 
of one state variable, hence only one state 
variable was measured while all others were 
unmeasured, and b) explicit dynamics, where 
the image pipeline was applied on charts of 
all state variables resulting in all 
measurable state variables. 
Table \ref{tbl:C3} compares the performance 
of EMMA and PySINDY. It shows while EMMA 
outperforms PySINDY on the parameter 
estimation task from charts, both EMMA and 
PySINDY performance decreases when the 
dynamics becomes implicit. However, EMMA has 
much less performance degradation than 
PySINDY.

\vspace{-8pt}
\begin{table}
	\centering
	\scriptsize
	\begin{tabular}{@{}p{0.35 in}@{}p{0.35 in}
	@{}p{0.6 in}@{}p{0.6 in}
	@{}p{0.6 in}@{}p{0.6 in}@{}}
	 \hline
		{\textbf{Example}} & { } 
		& \multicolumn{2}{c}{\textbf{EMMA}} 
		& \multicolumn{2}{c}{\textbf{PySINDY}} \\ \hline
   & & {Implicit} & {Explicit}
   & {Implicit} & {Explicit}\\ \hline
Lotka  & $\theta_{rmse}$ 
  & 0.054 $\pm$ 0.003
  & 0.048 $\pm$ 0.003 
  & 6.3 $\pm$ 1.7 
  & 0.054 $\pm$ 0.013 \\
Volterra & $x_{rmse}$ 
  & 0.03 $\pm$ 0.009
  & 0.03 $\pm$ 0.005 
  & 12.7 $\pm$ 2.4 
  & 0.05 $\pm$ 0.01 \\\hline
Chaotic  & $\theta_{rmse}$ 
  & 0.016$\pm$ 0.005 
  & 0.015 $\pm$ 0.005 
  & 2.3 $\pm$ 0.5 
  & 0.022 $\pm$ 0.009  \\
Lorenz & $x_{rmse}$ 
  & 1.7 $\pm$ 0.4 
  & 1.68 $\pm$ 0.4 
  & 37.4 $\pm$ 6.1 
  & 3.66 $\pm$ 1.1 \\\hline
 F8  & $\theta_{rmse}$ 
  & 7.81 $\pm$ 1.2 
  & 6.8 $\pm$ 1.7 
  & 21.9$\pm$ 4.2 
  & 10.5 $\pm$ 1.4 \\
Crusader& $x_{rmse}$ 
  & 1.6$\pm$ 0.3 
  & 1.57 $\pm$ 0.3 
  & 35.2 $\pm$ 7.3 
  & 3.46 $\pm$ 0.9 \\\hline
 HIV  & $\theta_{rmse}$ 
  & 0.45 $\pm$ 0.11 
  & 0.39 $\pm$ 0.09
  & 4.5 $\pm$ 0.9 
  & 0.43 $\pm$ 0.1 \\
therapy & $x_{rmse}$ 
  & 28.9$\pm$ 4.7 
  & 28.3 $\pm$ 3.2 
  & 89.1 $\pm$ 12.7 
  & 28.8 $\pm$ 2.7 \\\hline
 AID  & $\theta_{rmse}$ 
  & 0.51 $\pm$ 0.12 
  & 0.13 $\pm$ 0.04 
  & 6.5 $\pm$ 1.5 
  & 0.73 $\pm$ 0.2 \\
therapy & $x_{rmse}$ 
  & 8.7 $\pm$ 1.6 
  & 4.3 $\pm$ 1.7 
  & 79.6 $\pm$ 21.3
  & 31.2 $\pm$ 6.1  \\
		\bottomrule
	\end{tabular}
    \caption{Multi-Parameter extraction from 
    simulation charts. Comparison with and 
    without implicit dynamics}
    \label{tbl:C3}
\end{table}

\subsection{EMMA Execution Time}

\begin{table}[t]
\centering
\scriptsize
\resizebox{0.48\textwidth}{!}{
\begin{tabular}{lcccc}
\toprule
\textbf{Model} 
& \textbf{Avg Time/Epoch (s)} 
& \textbf{Std. Dev. (s)} 
& \textbf{\ Parameters} \\
\midrule
Delfys (Baseline) & 0.19 & 0.03 & 5.7M \\
EMMA (Ours)  & 0.37 & 0.01 & 53.2K \\
\bottomrule
\end{tabular}}
\caption{Training efficiency comparison 
between Delfys~\cite{garcia2025lfv} 
and EMMA.}
\label{tab:Exec}
\end{table}

Table \ref{tab:Exec} compares the execution 
time of EMMA with the next best performing 
comparator, Delfys~\cite{garcia2025lfv}. 
EMMA takes 1.4  $\times$ more time than  
Delfys on an NVIDIA RTX Ada 6000 GPU. 
This is because the LTC-NN 
requires solution of ordinary differential 
equation and is an inherently more 
computationally heavy operation. 
This overhead is offset by EMMA's novel 
capabilities (multi-parameter, forced, 
and implicit dynamics estimation) and 
its 107$\times$ smaller model size. 
EMMA's compact size also suits edge 
deployment; related model recovery 
architectures achieve 11$\times$ lower 
memory on FPGAs~\cite{merinda2024edge}.

\section{Conclusions}
\label{sec:conclusion}
\vspace{-6pt}
We introduced \textbf{EMMA}, a 
physics-informed multimodal framework 
that recovers dynamical parameters directly 
from raw video, audio, and images. By 
coupling an LTC-based estimator with 
invariant calibration and cross-modal 
alignment, EMMA infers both explicit system 
parameters and implicit dynamical components 
that are not directly sensed, then uses 
them to reproduce observed trajectories 
with high fidelity. The recovered parameters 
are interpretable and executable, enabling 
simulation, verification, and downstream 
control without bespoke postprocessing. 
Across diverse canonical systems and real 
platforms, EMMA delivers accurate parameter 
recovery with a compact pipeline that uses 
standard sensors and off-the-shelf tooling. 
We expect EMMA to provide a strong 
foundation for learning physical models 
from opportunistic multimodal data and for 
building physical AI agents.


\textbf{Limitations} include dependence on 
at least one temporally varying modality, 
a linear frequency-speed audio prior that 
may degrade under turbulence, sensitivity 
to severe camera shake, and higher runtime 
from LTC-based ODE integration.

\section*{Acknowledgments}
This project is partially funded by
DARPA AMP-N6600120C4020, 
DARPA FIRE-P000050426, NSF FDTBiotech grant
(2436801), NIH R21 grant (1R21HL175632).


{
    \small
    \bibliographystyle{ieeenat_fullname}
    \bibliography{main}
}


\clearpage
\twocolumn[{%
  \centering
  {\Large\textbf{EMMA: Extracting Multiple physical 
   parameters from Multimodal Data}\\[0.4em]}
  {\large Supplementary Material\par}
  \vspace{1.5em}
}]

\setcounter{section}{0}
\setcounter{table}{0}
\setcounter{figure}{0}
\setcounter{equation}{0}
\renewcommand\thesection{S\arabic{section}}
\renewcommand\thesubsection{\thesection.\arabic{subsection}}
\renewcommand\thetable{S\arabic{table}}
\renewcommand\thefigure{S\arabic{figure}}
\renewcommand\theequation{S\arabic{equation}}

\section{Ablation Study}
\label{app:ablations}

This supplement expands on the architecture 
behind EMMA's multi-modal, physics-informed 
estimator by isolating the impact of 
(i) forcing inputs and audio wavelength, 
(ii) the LTC network vs.\ alternative 
sequence models, (iii) implicit dynamics, 
and (iv) invariant knowledge. We follow the 
same dynamical systems introduced in the 
main paper (pendulum), and the real-world 
rover cases with hidden inputs. See the 
architecture and training details in the 
main paper (Fig.\ 2; Secs.\ 3-4). 
\emph{Where not stated otherwise, the loss 
and simulator are identical to the 
main setup.}

\subsection{Case Study with No Forcing 
Input (Pendulum)}
\label{app:abl_no_forcing}

\subsubsection{LTC Architecture: Pendulum}

\paragraph{Setup.} We ablated the LTC-NN 
using three alternative recurrent 
architectures (LSTM, GRU, and Transformer) 
as they share similar sequential structure. 
The pendulum example estimates two parameters 
and has no external force input $u(t)$, 
making it less complex.

\begin{table}[h]
  \centering
  \scriptsize
  \setlength{\tabcolsep}{4pt}

  \begin{tabular}{@{}lccc@{}}
    \toprule
    Architecture & 45\,cm & 90\,cm 
    & 150\,cm \\
    \midrule
    LSTM  & 0.49$\pm$0.20 
          & 0.98$\pm$0.41 
          & 1.35$\pm$0.69 \\
    GRU   & 0.49$\pm$0.20 
          & 0.98$\pm$0.41 
          & 1.35$\pm$0.69 \\
    Transformer & 0.49$\pm$0.20 
          & 1.07$\pm$0.42 
          & 1.35$\pm$0.69 \\
    \textbf{LTC} & 0.49$\pm$0.20 
          & 0.98$\pm$0.41 
          & 1.35$\pm$0.69 \\
    \midrule
    Ground Truth $L$ (m) & 0.45 
          & 0.90 & 1.50 \\
    \bottomrule
  \end{tabular}
  \caption*{(a) Estimated length $L$ (m)}

  \vspace{6pt}

  \begin{tabular}{@{}lccc@{}}
    \toprule
    Architecture & 45\,cm & 90\,cm 
    & 150\,cm \\
    \midrule
    LSTM  & 0.054$\pm$0.023 
          & 0.054$\pm$0.023 
          & 0.045$\pm$0.023 \\
    GRU   & 0.054$\pm$0.023 
          & 0.054$\pm$0.023 
          & 0.045$\pm$0.023 \\
    Transformer & 0.054$\pm$0.023 
          & 0.053$\pm$0.021 
          & 0.045$\pm$0.023 \\
    \textbf{LTC} & 0.054$\pm$0.023 
          & 0.054$\pm$0.023 
          & 0.045$\pm$0.023 \\
    Ground Truth & 0.05 
          & 0.05 & 0.05 \\
    \bottomrule
  \end{tabular}
  \caption*{(b) Damping time-constant 
  $\tau$ (1/s)}

  \vspace{3pt}
  \caption{Comparison of different 
  architectures using the Pendulum 
  (no forcing) example.}
  \label{tab:pendulum_no_u_transposed}
\end{table}

From the results in 
Table~\ref{tab:pendulum_no_u_transposed}, 
all architectures estimate very similar 
parameters with comparable accuracy, making 
our EMMA structure robust and versatile. 
This confirms that irrespective of 
architecture, EMMA is efficient on less 
complex examples with no force input.

\subsection{Case Study with Forcing Input}
\label{app:abl_forcing}

\subsubsection{LTC Architecture: Rover}
\label{app:abl_forcing_ltc}

\paragraph{Setup.} The motivation and 
justification for using LTC-NN arises when 
we have a forcing input $u(t)$ and a more 
complex example. We used the same three 
alternative architectures as for the 
pendulum but on the more complex Rover 
example, which has multiple parameters to 
estimate along with external force.

\begin{table}[h]
  \centering
  \scriptsize
  \begin{tabular}{lcccccc}
    \toprule
    Parameters & GT 
    & \textbf{LTC (Ours)} 
    & GRU & LSTM & Transformer \\
    \midrule
    X-arm length $m$ & 0.178 
      & \textbf{0.173} 
      & 0.202 & 0.168 & 0.169 \\
    Y-arm length $m$ & 0.144 
      & \textbf{0.133} 
      & 0.195 & 0.174 & 0.197 \\
    Mass $kg$ & 26.88 
      & \textbf{27.79} 
      & 39.50 & 39.45 & 39.28 \\
    CoM height $m$ & 0.112 
      & \textbf{0.119} 
      & 0.108 & 0.123 & 0.094 \\
    Wheel radius $r$ & 0.201 
      & \textbf{0.205} 
      & 0.196 & 0.212 & 0.188 \\
    \midrule
    Convergence epoch & -- 
      & 5 & 10 & 14 & 54 \\
    Training time/epoch & -- 
      & 63.57s & 22.36 
      & 25.07 & 62.82 \\
    \bottomrule
  \end{tabular}
  \caption{Comparison of different 
  architectures using the Rover 
  (forced dynamics).}
  \label{tab:rover_arch}
\end{table}

The results in 
Table~\ref{tab:rover_arch} show that the 
accuracy of other architectures degrades 
compared to LTC-NN on the more complex 
forced dynamics example. LTC-NN is therefore 
the most suitable architecture choice for 
EMMA, yielding accurate results with faster 
convergence.

\subsubsection{Multi-modal Ablation 
Without Audio}
\label{app:abl_forcing_modality}

\paragraph{Setup.} The use of multi-modal 
input is one of our key contributions, where 
we extract knowledge from different 
modalities making EMMA useful for various 
scenarios. We performed an ablation study 
on the rover example with video and audio 
input and compared it against video-only 
input. The setup was identical except the 
audio knowledge was removed.

\begin{table}[h]
  \centering
  \scriptsize
  \begin{tabular}{lcccc}
    \toprule
    Parameters & GT & Video+Audio 
    & Video-only \\
    \midrule
    X-arm length ($m$) & 0.178 
      & 0.163 & 0.189 \\
    Y-arm length ($m$) & 0.144 
      & 0.133 & 0.203 \\
    Mass ($kg$) & 26.88 
      & 27.79 & 39.64 \\
    CoM height ($m$) & 0.112 
      & 0.129 & 0.108 \\
    Wheel radius ($r$) & 0.201 
      & 0.245 & 0.171 \\
    \midrule
    Convergence epoch & -- & 5 & 30 \\
    Training time (s) & -- 
      & 54.1 & 122.3  \\
    \bottomrule
  \end{tabular}
  \caption{Effect of audio on parameter 
  recovery under forced dynamics.}
  \label{tab:modality_ablation}
\end{table}

Table~\ref{tab:modality_ablation} clearly 
demonstrates the importance of audio 
knowledge in parameter estimation. When 
more modalities are available, EMMA can 
observe and incorporate knowledge that 
better guides the model to estimate 
accurate parameters in less time.

\section{Physics Equations}
\label{app:equations}

We collect the governing equations for all 
systems used in the ablations, with 
parameters and units. These mirror the 
forms used in the simulator head of EMMA.

\subsection{Damped Pendulum}
For a pendulum with angle $\theta$, 
angular velocity $\omega = \dot{\theta}$, 
the dynamics follow:
\begin{align}
    \frac{d\theta}{dt} &= \omega. \\
    \frac{d\omega}{dt} &= 
    -\frac{g}{L}\sin\theta 
    - \frac{\tau}{L}\omega.
\end{align}
\noindent\textbf{Parameters:} 
$L \in (0.1, 2.0]$~m (length), 
$\tau \in (0, 0.5]$~s$^{-1}$ 
(damping coefficient), 
$g = 9.81$~m/s$^2$ (gravity).

\subsection{Torricelli Drainage}
For fluid draining through an orifice, 
height $h(t)$ evolves as:
\begin{equation}
    \frac{dh}{dt} = -K\sqrt{h}, \quad 
    K = C_d A_{\text{orifice}} 
    \sqrt{2g} / A_{\text{tank}}.
\end{equation}
\noindent\textbf{Parameters:} 
$K \in (0.001, 0.1]$ 
$\sqrt{\text{m}}/\text{s}$ 
(drainage coefficient).

\subsection{LED Exponential Decay}
Light intensity $I(t)$ follows 
first-order decay:
\begin{equation}
    \frac{dI}{dt} = -\gamma I(t), 
    \quad I(t) = I_0 e^{-\gamma t}.
\end{equation}
\noindent\textbf{Parameters:} 
$\gamma \in (0.01, 5.0]$~s$^{-1}$ 
(decay rate).

\subsection{Sliding Block with Friction}
Block on inclined plane with velocity $v$:
\begin{equation}
    \frac{dv}{dt} = g\sin(\alpha) 
    - \mu g\cos(\alpha).
\end{equation}
\noindent\textbf{Parameters:} 
$\alpha \in [10^\circ, 45^\circ]$ (incline), 
$\mu \in [0.1, 0.5]$ (friction).

\subsection{Free Fall}
Vertical velocity $v$ under 
quadratic drag:
\begin{equation}
    \frac{dv}{dt} = g - k v^2 
    \text{ sign}(v).
\end{equation}
\noindent\textbf{Parameters:} 
$k = \frac{C_d \rho A}{2m}$ 
(drag coefficient).

\subsection{Differential-Drive Rover 
(9 Parameters)}
The rover combines kinematic constraints 
with dynamic forces:
\begin{align}
    v &= \frac{r(\omega_r + \omega_l)}{2}, 
    \quad \dot{\psi} = 
    \frac{r(\omega_r - \omega_l)}{W}. \\
    m\dot{v}_x &= F_{\text{motor}} 
    - F_{\text{friction}} 
    - F_{\text{drag}}.
\end{align}

\noindent\textbf{Measured Parameters:} 
$a = 0.178$~m (X-arm), 
$b = 0.144$~m (Y-arm), 
$r = 0.201$~m (wheel radius), 
$m = 26.88$~kg, 
$W = 0.32$~m (wheelbase).

\noindent\textbf{Implicit Parameters:} 
$k_f = 0.15$ (friction), 
$C_d = 0.42$ (drag), 
$C_M = 0.112$~m (CoM height).

\subsection{6-DOF Quadrotor 
(12 Parameters)}
Full rigid-body dynamics with 
rotor dynamics:
\begin{align}
    \tau^2 \ddot{w}_i 
    + 2\zeta\tau\dot{w}_i 
    + w_i &= k_p u_i. \\
    T_i = k_{Th} w_i^2, 
    \quad &\tau_i = k_{To} w_i^2. \\
    m\ddot{\mathbf{p}} &= 
    R(\mathbf{q})\mathbf{T} 
    - m g\mathbf{e}_z 
    - \mathbf{F}_{\text{drag}}.
\end{align}

\noindent\textbf{Measured Parameters:} 
$k_{Th} = 1.1 \times 10^{-5}$ 
N$\cdot$s$^2$/rad$^2$, 
$k_{To} = 1.3 \times 10^{-7}$ 
N$\cdot$m$\cdot$s$^2$/rad$^2$, 
$d_{xm} = 0.18$~m, 
$d_{ym} = 0.20$~m, 
$d_{zm} = 0.07$~m.

\noindent\textbf{Audio-Inferred:} 
$k_p = 0.91$, $\tau = 0.012$~s, 
$\zeta = 0.7$.

\section{Differentiable Trajectory Rollout}
\label{app:simulation}

To ensure numerical stability and consistency 
with the architecture layout in Fig.~2 of 
the main paper, we employ a differentiable 
4th-order Runge-Kutta (RK4) integrator. This 
provides higher-order error control compared 
to standard Euler integration, which is 
critical for stiff dynamical systems like 
the quadrotor.

Given estimated parameters 
$\hat{\boldsymbol{\theta}}$ from the LTC 
network and the continuous physics function 
$\mathbf{f}$, the state update from time 
$t$ to $t+1$ is computed as:

\begin{align}
    k_1 &= f(x_t, u_t; \hat{\theta}). \\
    k_2 &= f\left(x_t 
    + \frac{\Delta t}{2}k_1, 
    u_{t+\frac{1}{2}}; 
    \hat{\theta}\right). \\
    k_3 &= f\left(x_t 
    + \frac{\Delta t}{2}k_2, 
    u_{t+\frac{1}{2}}; 
    \hat{\theta}\right). \\
    k_4 &= f(x_t + \Delta t k_3, 
    u_{t+1}; \hat{\theta}). \\
    x_{t+1} &= x_t + \frac{\Delta t}{6} 
    (k_1 + 2k_2 + 2k_3 + k_4).
\end{align}

where the simulation time step is clamped at 
$\Delta t = \min(0.03, \text{fps}^{-1})$. 
Intermediate control inputs 
$\mathbf{u}_{t+\frac{1}{2}}$ are obtained 
via linear interpolation of the forcing 
signal. The simulation runs for 
$T_{\text{sim}} = \min(500, T)$ steps. 
Parameter physical constraints are enforced 
via soft clamping: 
$\theta_i \gets \max(\epsilon, \theta_i)$ 
with $\epsilon = 10^{-4}$.

\section{Additional Robustness 
  and Ablation Experiments}
\label{app:rebuttal_experiments}

This section reports five additional 
experiments validating EMMA's robustness 
across feature extraction backbones, 
architecture choices, initialization 
sensitivity, statistical reproducibility, 
and audio noise levels.

\subsection{Optical Flow vs.\ YOLO 
  (Detector Agnosticism)}
\label{app:optical_flow}

To verify that EMMA's physics extraction 
is independent of the object detection 
front-end, we replaced YOLOv11 with 
unsupervised Farneback optical flow 
tracking on both the rover and pendulum 
systems. 
Table~\ref{tab:supp_yolo_optflow} 
shows that optical flow achieves 
comparable accuracy, confirming that 
EMMA's core contribution lies in the 
LTC physics layer rather than the 
feature extractor.

\begin{table}[h]
\centering
\scriptsize
\setlength{\tabcolsep}{5pt}
\renewcommand{\arraystretch}{1.15}

\begin{minipage}[t]{0.48\textwidth}
\centering
\textbf{(a) Rover}\par\vspace{3pt}
\begin{tabular}{@{}lccc@{}}
\toprule
Parameter & Ground Truth 
  & YOLOv11 
  & Optical Flow \\
\midrule
$a$ (m) & 0.178 
  & 0.196 & \textbf{0.184} \\
$b$ (m) & 0.144 
  & \textbf{0.134} & 0.110 \\
$r$ (m) & 0.201 
  & 0.223 & \textbf{0.202} \\
\bottomrule
\end{tabular}
\end{minipage}\hfill
\begin{minipage}[t]{0.48\textwidth}
\centering
\textbf{(b) Pendulum ($L$, m)}%
\par\vspace{3pt}
\begin{tabular}{@{}lccc@{}}
\toprule
Configuration & Ground Truth 
  & YOLOv11 
  & Optical Flow \\
\midrule
45\,cm & 0.45 
  & \textbf{0.50$\pm$0.04} 
  & 0.66$\pm$0.00 \\
90\,cm & 0.90 
  & 0.86$\pm$0.07 
  & \textbf{0.89$\pm$0.00} \\
150\,cm & 1.50 
  & \textbf{1.50$\pm$0.00} 
  & 1.49$\pm$0.00 \\
\bottomrule
\end{tabular}
\end{minipage}

\vspace{4pt}
\caption{YOLOv11 vs.\ unsupervised 
  optical flow on rover and pendulum. 
  Optical flow achieves comparable 
  accuracy without any pretrained 
  detector. Best per row in bold.}
\label{tab:supp_yolo_optflow}
\end{table}

\subsection{Statistical Validation 
  (Multi-Seed Reproducibility)}
\label{app:stat_validation}

Table~\ref{tab:supp_statistical} reports 
rover parameter estimates over 5 random 
seeds (42--46), providing standard 
deviations that quantify 
reproducibility. Average error across four 
measurable parameters is 
9.5\% $\pm$ 8.9\%.

\begin{table}[h]
\centering
\scriptsize
\setlength{\tabcolsep}{6pt}
\renewcommand{\arraystretch}{1.15}
\begin{tabular}{@{}lccr@{}}
\toprule
Parameter 
  & Ground Truth 
  & Mean $\pm$ Std 
  & Error (\%) \\
\midrule
$a$ (m) & 0.178 
  & 0.184 $\pm$ 0.020 & 3.4 \\
$b$ (m) & 0.144 
  & 0.176 $\pm$ 0.031 & 22.2 \\
$r$ (m) & 0.201 
  & 0.222 $\pm$ 0.027 & 10.4 \\
$CM$ (m) & 0.112 
  & 0.114 $\pm$ 0.008 & 1.8 \\
\bottomrule
\end{tabular}
\caption{Rover statistical validation 
  over 5 random seeds (42--46). 
  $\pm$ denotes standard deviation 
  across seeds.}
\label{tab:supp_statistical}
\end{table}

\subsection{Drone Statistical Validation 
  (Multi-Seed Reproducibility)}
\label{app:drone_stat_validation}

Table~\ref{tab:supp_drone_statistical} 
reports drone parameter estimates over 
5 random seeds (42--46), providing 
standard deviations that quantify 
reproducibility. Average error across 
all measurable parameters is 
$17.5\%$.

\begin{table}[h]
\centering
\scriptsize
\setlength{\tabcolsep}{6pt}
\renewcommand{\arraystretch}{1.15}
\begin{tabular}{@{}lccr@{}}
\toprule
\textbf{Parameter} 
  & \textbf{Ground Truth} 
  & \textbf{Mean $\pm$ Std} 
  & \textbf{Error (\%)} \\
\midrule
$k_{Th}$  & 1.1   
  & 1.076 $\pm$ 0.000 & 2.2 \\
$k_{To}$  & 1.3   
  & 1.632 $\pm$ 0.000 & 25.5 \\
$k_p$     & 0.91  
  & 1.000 $\pm$ 0.000 & 9.9 \\
$\tau_2$  & 0.012 
  & 0.015 $\pm$ 0.000 & 25.0 \\
$d_{xm}$  & 0.18  
  & 0.160 $\pm$ 0.000 & 11.1 \\
$d_{ym}$  & 0.20  
  & 0.160 $\pm$ 0.000 & 20.0 \\
$d_{zm}$  & 0.07  
  & 0.050 $\pm$ 0.000 & 28.6 \\
\bottomrule
\end{tabular}
\caption{Drone statistical validation 
  over 5 random seeds (42--46). 
  $\pm$ denotes standard deviation 
  across seeds.}
\label{tab:supp_drone_statistical}
\end{table}

\subsection{LTC vs.\ Neural ODE 
  vs.\ CT-GRU}
\label{app:arch_comparison}

We compare LTC against two 
continuous-time alternatives: Neural ODE and CT-GRU. 
Table~\ref{tab:supp_arch} shows that 
all architectures perform comparably 
on the unforced pendulum. Under forcing 
inputs (rover), LTC outperforms Neural ODE by 
approximately 25\% and CT-GRU 
by approximately 5\% in 
average parameter error, validating 
that input-dependent time constants 
are critical for modeling forced 
dynamics.

\begin{table}[h]
\centering
\scriptsize
\setlength{\tabcolsep}{5pt}
\renewcommand{\arraystretch}{1.15}
\begin{minipage}[t]{0.48\textwidth}
\centering
\textbf{(a) Rover (Forcing Input)}%
\par\vspace{3pt}
\begin{tabular}{@{}lcccc@{}}
\toprule
Parameter & Ground Truth 
  & LTC & Neural ODE 
  & CT-GRU \\
\midrule
$a$ (m) & 0.178 
  & 0.196 & 0.238 
  & \textbf{0.179} \\
$b$ (m) & 0.144 
  & \textbf{0.134} 
  & 0.212 & 0.186 \\
$r$ (m) & 0.201 
  & \textbf{0.223} 
  & 0.244 & 0.226 \\
\midrule
Avg.\ Error & 
  & \textbf{9.3\%} 
  & 34.1\% & 14.1\% \\
\bottomrule
\end{tabular}
\end{minipage}\hfill
\begin{minipage}[t]{0.48\textwidth}
\centering
\textbf{(b) Pendulum (No Forcing)}%
\par\vspace{3pt}
\begin{tabular}{@{}lcccc@{}}
\toprule
Configuration & Ground Truth 
  & LTC & Neural ODE 
  & CT-GRU \\
\midrule
45\,cm & 0.45 
  & \textbf{0.50} 
  & 0.59 & 0.54 \\
90\,cm & 0.90 
  & 0.86 
  & 0.99 
  & \textbf{0.94} \\
150\,cm & 1.50 
  & \textbf{1.50} 
  & 1.56 & 1.48 \\
\midrule
Avg.\ Error & 
  & \textbf{5.2\%} 
  & 15.0\% & 8.6\% \\
\bottomrule
\end{tabular}
\end{minipage}
\vspace{4pt}
\caption{Continuous-time architecture 
  comparison. All architectures perform 
  comparably without forcing; under 
  forcing inputs, LTC outperforms 
  Neural ODE by approximately 25\% 
  and CT-GRU by approximately 5\%. 
  Best per row in bold.}
\label{tab:supp_arch}
\end{table}

\subsection{Initialization Sensitivity}
\label{app:init_sensitivity}

To evaluate robustness to poor 
initialization, we expand parameter 
bounds by 200\% and initialize far 
from ground truth. 
Table~\ref{tab:supp_init} shows EMMA 
achieves $<$10\% error in 5 out of 6 
configurations, confirming that 
accurate estimation does not require 
initialization close to ground truth.

\begin{table}[h]
\centering
\scriptsize
\setlength{\tabcolsep}{5pt}
\renewcommand{\arraystretch}{1.15}

\begin{minipage}[t]{0.48\textwidth}
\centering
\textbf{(a) Sliding Block ($\alpha$)}%
\par\vspace{3pt}
\begin{tabular}{@{}lcccc@{}}
\toprule
Config & GT 
  & Init 
  & Est 
  & Error (\%) \\
\midrule
Low & 20$^\circ$ 
  & 25$^\circ$ 
  & 20.69$^\circ$ 
  & \textbf{3.45} \\
Mid & 25$^\circ$ 
  & 30$^\circ$ 
  & 25.48$^\circ$ 
  & \textbf{1.92} \\
High & 30$^\circ$ 
  & 45$^\circ$ 
  & 28.25$^\circ$ 
  & \textbf{5.84} \\
\bottomrule
\end{tabular}
\end{minipage}\hfill
\begin{minipage}[t]{0.48\textwidth}
\centering
\textbf{(b) Pendulum ($L$, m)}%
\par\vspace{3pt}
\begin{tabular}{@{}lcccc@{}}
\toprule
Config & GT 
  & Init 
  & Est 
  & Error (\%) \\
\midrule
45\,cm & 0.45 
  & 0.85 & 0.447 
  & \textbf{0.73} \\
90\,cm & 0.90 
  & 1.30 & 0.683 
  & 24.08 \\
150\,cm & 1.50 
  & 1.10 & 1.620 
  & \textbf{8.02} \\
\bottomrule
\end{tabular}
\end{minipage}

\vspace{4pt}
\caption{Initialization sensitivity 
  with 200\% expanded bounds and 
  distant initialization. Bold 
  indicates $<$10\% error. EMMA 
  converges accurately in 5 of 6 
  configurations.}
\label{tab:supp_init}
\end{table}

\subsection{Audio Noise Robustness}
\label{app:audio_noise}

We inject additive Gaussian noise at 
SNR levels of 20, 10, and 5\,dB into 
the rover audio stream. 
Table~\ref{tab:supp_audio} shows that 
parameter estimates vary by less than 
1.1\% across all noise levels, 
demonstrating that EMMA's audio 
pipeline degrades gracefully under 
realistic acoustic interference.

\begin{table}[h]
\centering
\scriptsize
\setlength{\tabcolsep}{6pt}
\renewcommand{\arraystretch}{1.15}
\begin{tabular}{@{}lccccc@{}}
\toprule
Parameter 
  & Ground Truth 
  & SNR 20\,dB 
  & SNR 10\,dB 
  & SNR 5\,dB 
  & Var (\%) \\
\midrule
$a$ (m) & 0.178 
  & 0.205 & 0.205 & 0.205 
  & 0.15 \\
$b$ (m) & 0.144 
  & 0.165 & 0.165 & 0.165 
  & 0.55 \\
$r$ (m) & 0.201 
  & 0.184 & 0.186 & 0.186 
  & 1.08 \\
$CM$ (m) & 0.112 
  & 0.115 & 0.115 & 0.115 
  & 0.04 \\
\bottomrule
\end{tabular}
\caption{Rover audio noise robustness. 
  Additive Gaussian noise at three SNR 
  levels causes $<$1.1\% variation in 
  all estimated parameters.}
\label{tab:supp_audio}
\end{table}

\end{document}